\documentclass{article}

 \usepackage[final]{corl_2019}
\usepackage{amsmath}
\usepackage{graphicx,subfigure}
\usepackage[linesnumbered, ruled]{algorithm2e}
\usepackage{floatrow}
\usepackage{multicol}
\usepackage{multirow}
\usepackage{caption}
\newfloatcommand{capbtabbox}{table}[][\FBwidth]
\usepackage{enumitem}
\usepackage{bm}
\begin{document}
%\ShortHeadings{Probabilistic Trajectory Mapping}
%\firstpageno{1}

%\title{Kernel Trajectory Maps: Probabilistic Functional Trajectory Learning in Dynamic Environments}

\title{Kernel Trajectory Maps for Multi-Modal Probabilistic Motion Prediction}

\author{Weiming Zhi $^{1}$ and
        %Department of Computer Science,\\ University of Sydney, Australia\\
        %\And
        Lionel Ott $^{1}$ and
        %Department of Computer Science,\\ University of Sydney, Australia\\
        %\And
        Fabio Ramos $^{1,2}$ \\
       $^{1}$ Department of Computer Science, University of Sydney, Australia\\
       $^{2}$ NVIDIA, USA \\
       \texttt{\{firstname.lastname\}@sydney.edu.au}\\}

\maketitle

\begin{abstract}   
Understanding the dynamics of an environment, such as the movement of humans and vehicles, is crucial for agents to achieve long-term autonomy in urban environments. This requires the development of methods to capture the multi-modal and probabilistic nature of motion patterns. We present \textbf{k}ernel \textbf{t}rajectory \textbf{m}aps (KTM) to capture the trajectories of movement in an environment. KTMs leverage the expressiveness of kernels from non-parametric modelling by projecting input trajectories onto a set of representative trajectories, to condition on a sequence of observed waypoint coordinates, and predict a multi-modal distribution over possible future trajectories. The output is a mixture of continuous stochastic processes, where each realisation is a continuous functional trajectory, which can be queried at arbitrarily fine time steps.

%then a single hidden layer neural network can be used to effectively learn a mixture of stochastic processes. Conditioned on a sequence of observed waypoints of arbitrary length, KTMs are able to provide multi-modal probability distributions over future trajectories. These distributions can then be sampled to generate possible motion trajectories. KTMs models the global location of provided waypoints, and the full history of the trajectory. Each sample from the distribution of trajectories is a functional continuous trajectory, allowing us to query the output at arbitrarily fine time resolutions. 
\end{abstract}

\section{Introduction}
Autonomous agents may be required to operate in environments with moving objects, such as pedestrians and vehicles in urban areas, for extended periods of time. A probabilistic model that captures the movement of surrounding dynamic objects allows an agent to make more effective and robust plans. This work presents \textbf{k}ernel \textbf{t}rajectory \textbf{m}aps (KTM) \footnote{Code available at \url{https://github.com/wzhi/KernelTrajectoryMaps}}, that capture the multi-modal, probabilistic, and continuous nature of future paths. Given a sequence of observed waypoints of a trajectory up to a given coordinate, a KTM is able to produce a multi-modal distribution over possible future trajectories, represented by a mixture of stochastic processes. Continuous functional trajectories, which are functions mapping queried times to trajectory coordinates, can then be sampled from the output stochastic process.

Early methods to predict future motion trajectories generally extrapolate based on physical laws of motion \citep{SurveyDynamics}. Although simple and often effective, these models have the drawback of being unable to make use of other observed trajectories, or account for environment topology. For example, physics-based methods fail to take into account that trajectories may follow a road that exists in a map. To address this shortcoming, methods have been developed that map the direction or flow of movements in an environment in a probabilistic manner \citep{sptempFlow, DirectionalGridMaps, Flow, KernelEmbeddingDirections}. These methods are able to output distributions over future movement directions or velocities, conditioned on the current queried coordinate. Using these models, one can sequentially forward sample to obtain a trajectory. This forward sampling approach makes the Markov assumption, assuming that the object dynamics only depend on the current position of the object. These approaches discard useful information from the trajectory history, and can accumulate errors from the forward simulation.

\begin{figure}
    \centering
    \begin{subfigure}
     \centering
     \includegraphics[width=0.24\textwidth]{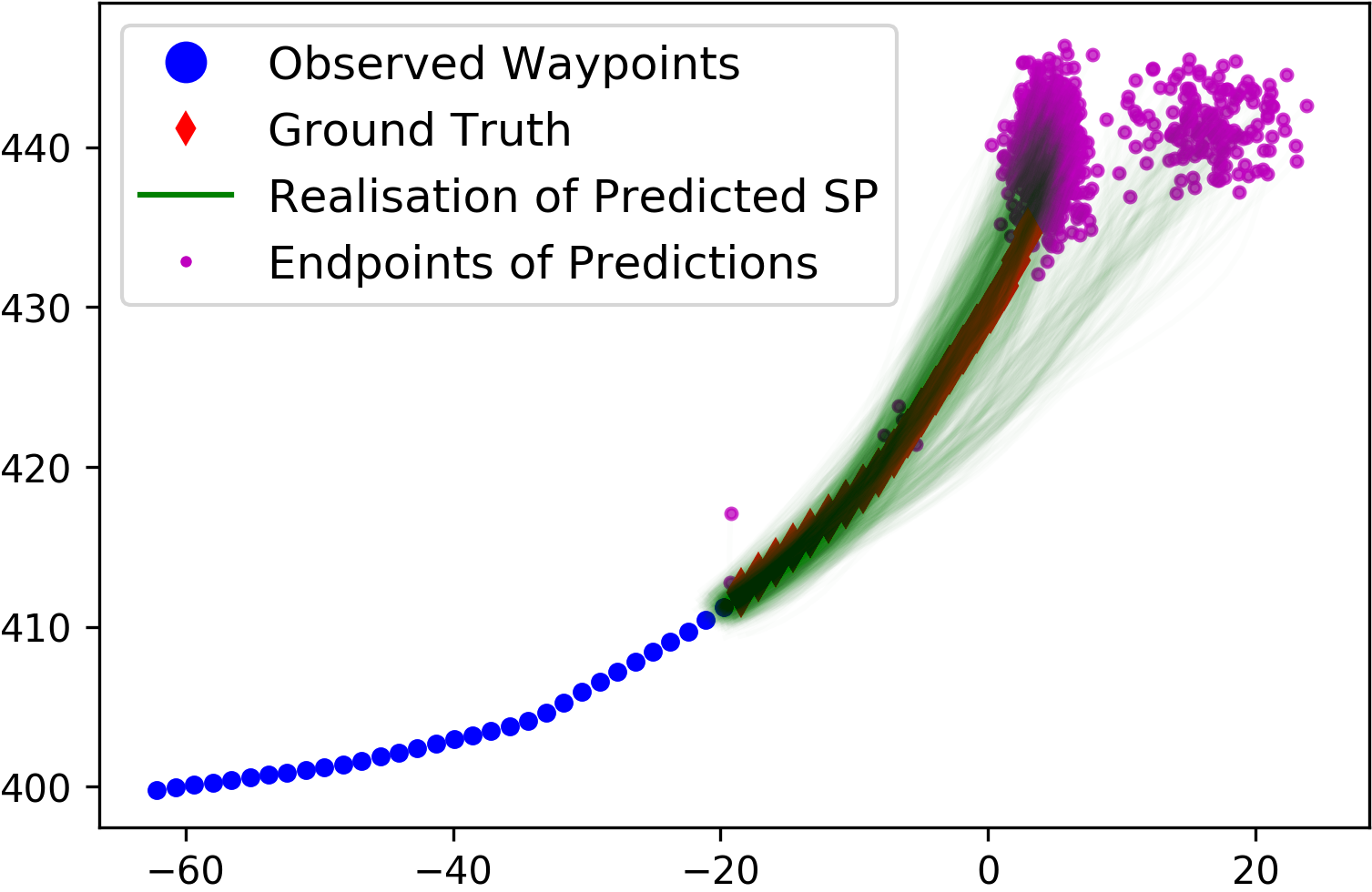}
    \end{subfigure}
     %\hspace{0.1\textwidth}
    \begin{subfigure}
     \centering
     \includegraphics[width=0.24\textwidth]{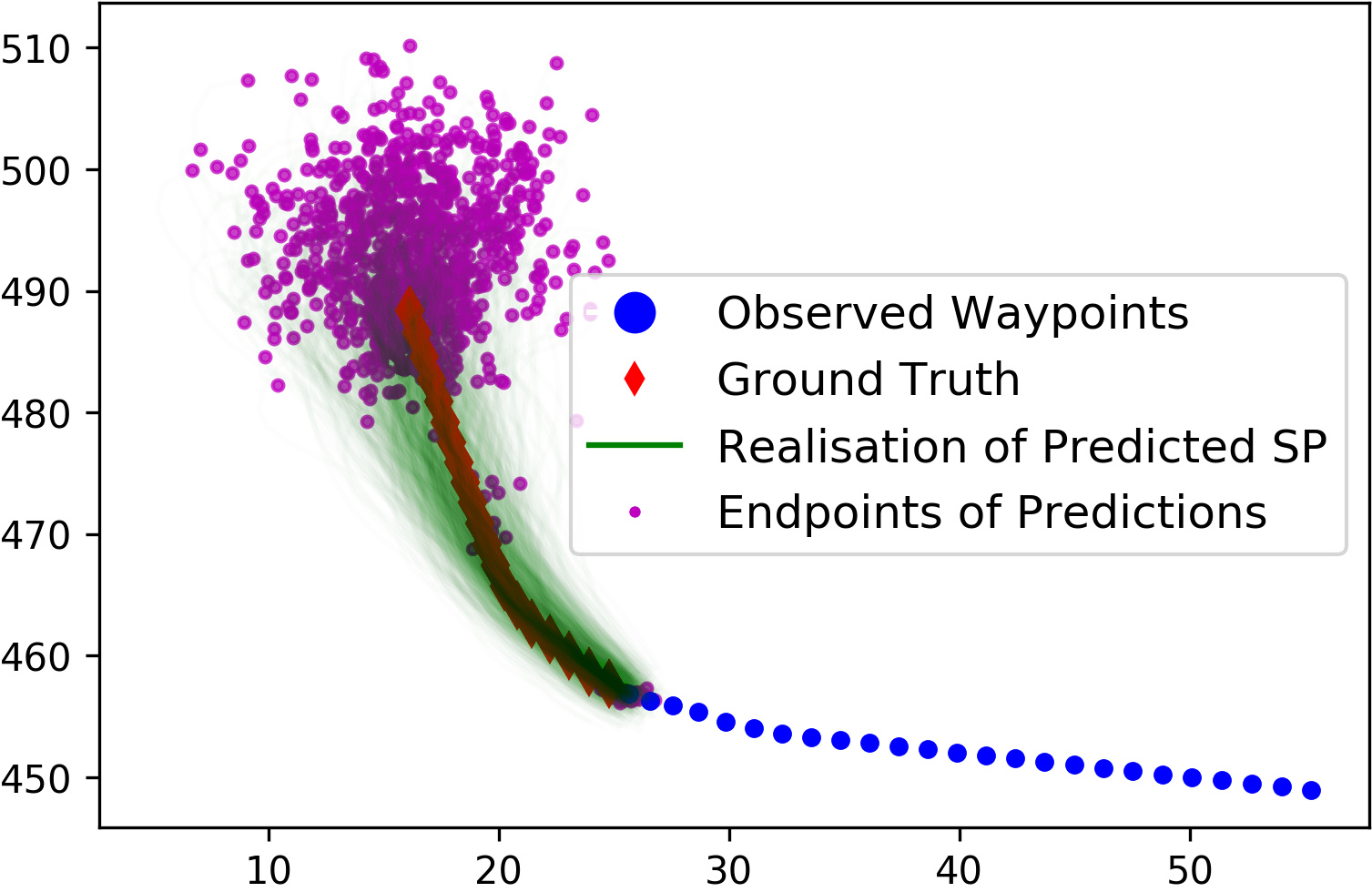}
    \end{subfigure}
    \begin{subfigure}
     \centering
     \includegraphics[width=0.24\textwidth]{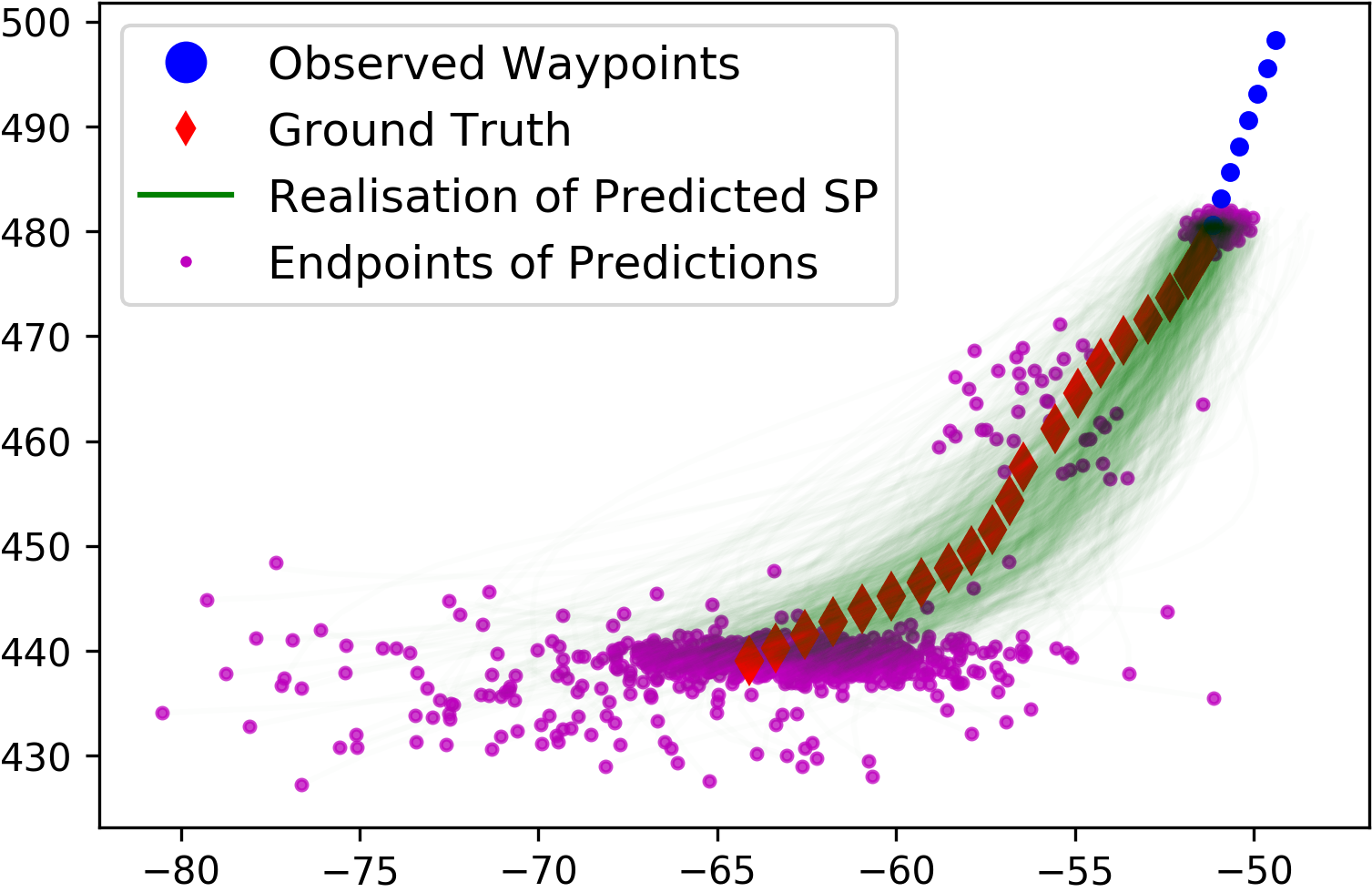}
    \end{subfigure}
    \begin{subfigure}
     \centering
     \includegraphics[width=0.24\textwidth]{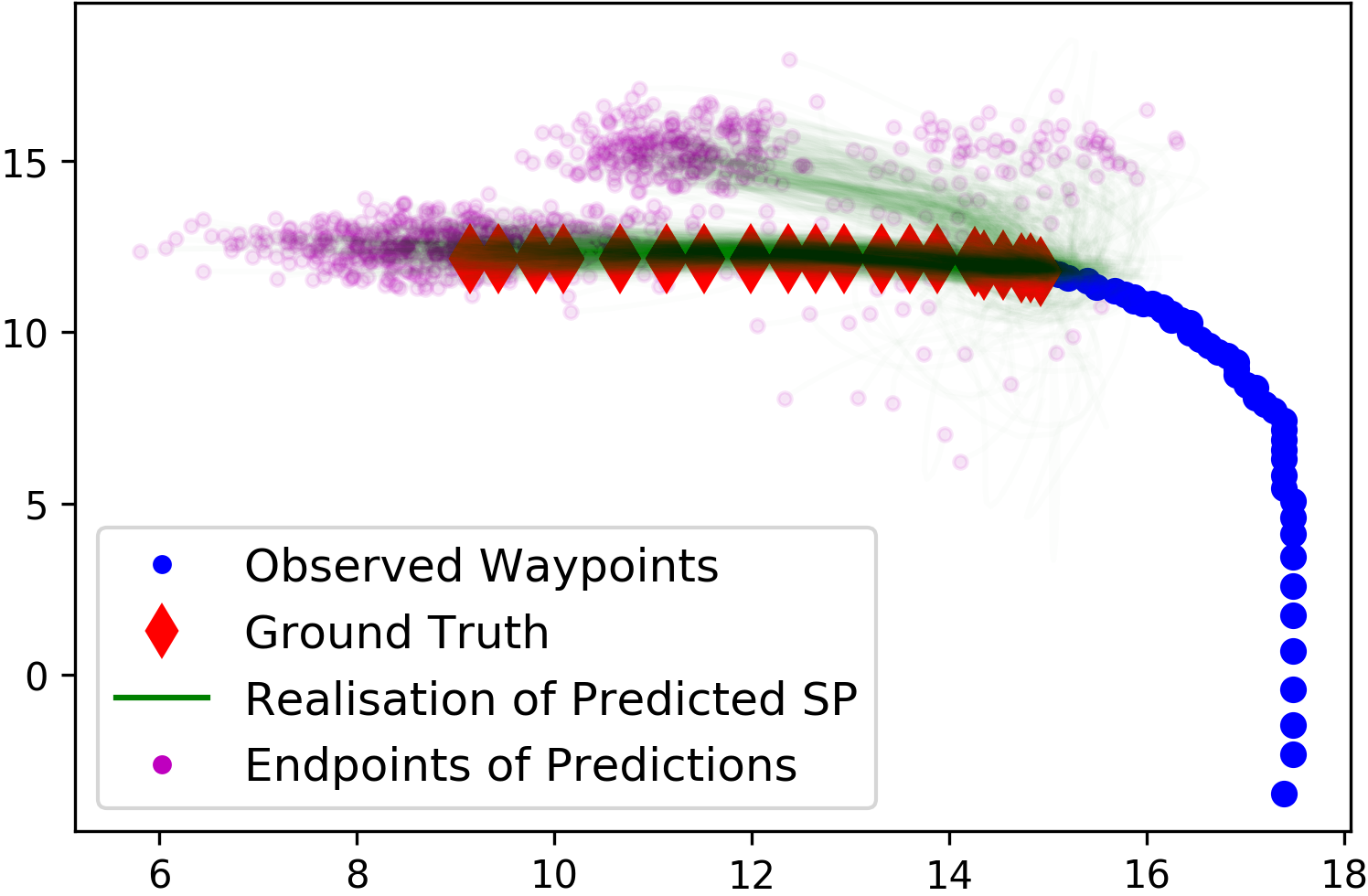}
    \end{subfigure}
    \vspace*{-3mm}
    \caption{Observed waypoints (blue) and predicted trajectories (green with magenta end-points) sampled from KTM outputs. The ground truth is indicated in red. The probabilistic and multi-modal nature of KTMs is able to capture the complexity of the motion patterns. Units in meters.}\label{ExamplePlots1}
\end{figure}

Motivated to overcome these aforementioned limitations of past methods, we utilise \emph{distance substitution kernels} \citep{DistanceKernel,NipsDistanceKernel} with the Fr\'echet distance \citep{OriginalFrechet, DiscreteFrechet, TrajectoryDistances} to project trajectory data onto a representative set of trajectories to obtain high-dimensional projection features. Using a neural network with a single hidden layer with the projection features, we learn a multi-modal mixture of stochastic processes. The resulting mixture is also a stochastic process, and can be viewed as a distribution over functions, where each realisation is a continuous functional trajectory. Figure \ref{ExamplePlots1} shows observed trajectories and realised trajectory predictions, demonstrating the probabilistic and multi-modal nature of KTMs. The probabilistic nature of the output provides an estimate for uncertainty, which can be used for robust planning and decision making. We contribute the KTM, a method that:
\begin{enumerate}
    \item is trajectory history aware and captures dependencies over the entire trajectory;
    \item models the output as a mixture of stochastic process, providing a multi-modal distribution over possible trajectories;
    \item represents realised trajectories as continuous functions, allowing them to be queried at arbitrary time resolution. 
\end{enumerate}

\section{Related Work}
Kernel Trajectory Maps (KTMs) learn motion patterns in an environment, and represent sampled outputs as continuous trajectories. i.e. trajectories that can be queried at arbitrarily fine time resolutions. Here we briefly revisit literature on modelling motion dynamics and continuous trajectories.

\textbf{Motion Modelling}. Some of the simplest approaches to model trajectory patterns are kinematic models that make extrapolations based on a sequence of observed coordinates. Popular examples include the constant velocity and constant acceleration models \citep{HumanMotionSurvey}. Some other attempts to understand dynamics take the approach of extending occupancy mapping beyond static environments by building occupancy representations along time \citep{TOG1,TOG2,TOG3}. This approach tends to be memory intensive, limiting scalability. Other recent approaches have incorporated global spatial \citep{sptempFlow,DirectionalGridMaps,Flow, KernelEmbeddingDirections} and temporal information \citep{sptempFlow, FreMen,Molina2018}. The authors of \citep{DirectionalGridMaps} propose \emph{directional grid maps}, a model that learns the distribution of motion directions in each grid cell of a discretised environment. This is achieved by fitting a mixture of von-Mises distributions on the motion directions attributed to each cell. A similar method is also presented in \citep{Flow}, where a map of velocity distributions in the environment is modelled by semi-wrapped Gaussian mixture models. Continuous spatiotemporal extensions are provided in \citep{sptempFlow}. These methods are able to capture the uncertainty of motion at a given point coordinate, but require forward sampling to obtain trajectories.

\textbf{Continuous Trajectories}.
Continuous representations of trajectories, often modelled by a Gaussian processes \citep{Rasmussen2004} or a sparse low rank approximations of Gaussian processes \citep{SparseGP}, have arisen in previous works for trajectory estimation \citep{barfootTrajEst} and motion planning \citep{FunctionalMotionPlanning1, FunctionalMotionPlanning2}. In this work, we also formulate a method to produce continuous trajectories, and then leverage continuous trajectories for extrapolation, rather than the estimation and interpolation problems addressed in previous works. 

\section{Methodology}
\subsection{Problem Formulation and Overview}
We work with continuous trajectory outputs, $\bm{\Xi}$, and discrete trajectories inputs, $\bm{\xi}$. Discrete trajectories are an ordered set of waypoint coordinates indexed by time, $\bm{\xi}=\{(x_t,y_t)\}^{T}_{t=1}$. Continuous trajectories, $\bm{\Xi}(\cdot)$, are functions that map time to coordinates. Continuous trajectories can be discretised by querying at time steps, $\bm{t}=1,\ldots,T$. In this paper, continuous trajectories, $\bm{\Xi(\cdot)}$, are defined by weighted combinations of features, $\bm{\phi(\cdot)}$, where $\bm{w}$ contains the weight parameters. $\bm{\phi(\cdot)}$ is dependent on the queried time. We discuss continuous trajectories in detail in subsection \ref{functionalTraj}. 

Given a dataset of $N$ pairs of trajectories, $\mathcal{D}=\{\bm{\xi}^{\text{Obs}}_n, \bm{\xi}^{\text{Tar}}_n\}^{N}_{n=1}$, where $\bm{\xi}^{\text{Obs}}$ is an observed input trajectory, and $\bm{\xi}^{\text{Tar}}$ is a target trajectory. The input contains coordinates up to a given time, and the target is a continuation of the same trajectory thereafter. We seek to predict a probability distribution over possible future trajectories beyond the given observed waypoints, $p\big(\bm{\Xi}^{*}(\cdot)|\bm{\xi}^{*},\mathcal{D}, \bm{\phi(\bm{\cdot})}\big),
$
where $\bm{\xi}^{*}$ is a queried discrete trajectory, $\bm{\Xi}^{*}(\cdot)$ is a predicted continuous trajectory starting from the last time step of $\bm{\xi}^{*}$. To find the distribution over future trajectories, we write the marginal likelihood as,
\begin{equation}\label{MarginaliseEq}
    p(\bm{\Xi}^{*}(\cdot)|\bm{\xi}^{*},\mathcal{D}, \bm{\phi})=\int p(\bm{\Xi}^{*}(\cdot)|\bm{\phi}, \bm{w})p(\bm{w}|\bm{\xi}^{*},\mathcal{D})d\bm{w}.
\end{equation}
To evaluate the marginal likelihood, we learn $p\big(\bm{w}|\bm{\xi}^{*},\mathcal{D}\big)$ and sample realisations of weights to conduct inference (detailed in subsection \ref{inferenceSubsection}). This learning can be summarised by the following steps:

\begin{enumerate}
    \item Construct high-dimensional feature vectors of observed discrete trajectories, by projecting to a set of representative trajectories, using discrete Fr\'echet \citep{DiscreteFrechet} kernels (DF-Kernels). (Subsection \ref{Subsection:genfeat})
    
    \item Concisely represent each trajectory as a continuous function, defined by a vector of weights and predetermined basis functions. (Subsection \ref{functionalTraj})
    
    \item Train a single hidden layer mixture density network (MDN) model on the projection features, with weight vectors as targets, to obtain $p\big(\bm{w}|\bm{\xi}^{*},\mathcal{D}\big)$. (Subsection \ref{Learning Mixture})
    
\end{enumerate}
%We shall consider a dataset of $N$ pairs of observed trajectories, $\mathcal{D}=\{\bm{\xi}^{\text{Obs}}_n, \bm{\xi^{\text{Tar}}}_n\}^{N}_{n=1}$, where, at the $n_{th}$ data entry, $\bm{\xi^x}_n$ is an input trajectory, and $\bm{\xi^y}_n$ is a target trajectory we wish to learn. Both $\bm{\xi^x}_n$ and $\bm{\xi^y}_n$ are ordered sets of coordinates sampled at fixed time intervals, with $\bm{\xi^{\text{Obs}}}_n$ containing the coordinates from time $t=1\ldots T'_{n}$ and $\bm{\xi^{\text{Tar}}}_n$ the coordinates from times $t=T'_n, \ldots, T_n$. $T'$ is the last time step of observed waypoints, and the beginning of trajectory predictions. For the $nth$ observation, we denote the trajectories as $\bm{\xi^{\text{Obs}}}_n=\{(x_t,y_t)\}^{T'_n}_{t=1}$ and $\bm{\xi^{\text{Tar}}}_n=\{(x_t,y_t)\}^{T_n}_{t=T'_n}$. We seek to predict a probability distribution over possible future trajectories given observed waypoints, 

%The projections over the trajectories encode the relative similarity between sequences. This enables conditioning on ordered sets of coordinates in space, extending beyond solely considering single coordinates. While the mixture density model provides an output of a mixture of stochastic processes, each component of this mixture can have a different mean function, permitting the output to have several modes.

\subsection{Generating Projection Features from Discrete Trajectories}\label{Subsection:genfeat}
In this subsection, we describe the conversion from discrete input trajectories to high-dimensional kernel projections. We make use of distance substitute kernels \citep{DistanceKernelsICDM, DistanceKernel, NipsDistanceKernel}, which are defined as $k(x,x')=k(d(x,x'))$, for kernel function, $k(\cdot)$, and distance measure $d$ that is symmetric, i.e. $d(x,x')=d(x',x)$, and has zero diagonal, i.e. $d(x,x)=0$. In this work, we use the discrete Fr\'echet distance \citep{DiscreteFrechet} substituted in a radial basis function (RBF) kernel. The Fr\'{e}chet distance \citep{OriginalFrechet} between curves is defined as,
\begin{equation}
    Fr(P,Q)=\inf_{\alpha,\beta}\max_{t\in[0,1]}||P(\alpha(t))-Q(\beta(t))||,
\end{equation}
where $P,Q$ are parameterisations of two curves, and $\alpha, \beta$ range over all continuous and monotone increasing functions. We use a discrete approximation of the Fr\'echet distance, which provides a distance metric between ordered sets of arbitrary length. The discrete Fr\'echet distance between two trajectories can be computed efficiently in $O(pq)$, where there are $p$ and $q$ waypoints in each of the trajectories. The discrete Fr\'echet distance takes into consideration the ordering of waypoints, and can in general distinguish a given trajectory with its reverse. An algorithm to compute the discrete Fr\'echet distance is outlined in \citep{DiscreteFrechet}. We name this kernel the \emph{discrete Fr\'echet (DF) kernel}, given by:
\begin{equation}\label{DFKernel}
    k_{DF}(\bm{\xi},\bm{\xi'})=\exp\bigg\{-\frac{\big(d_{DF}(\bm{\xi},\bm{\xi'})\big)^{2}}{2\ell_{DF}}\bigg\},
\end{equation}
where $\bm{\xi}$ and $\bm{\xi'}$ are discrete trajectories, which can be of different lengths; $\ell_{DF}$ is the length scale parameter of the RBF kernel; $d_{DF}$ is the discrete Fr\'{e}chet distance.

%The Fr\'{e}chet distance, $d_{F}$, is defined for curves $f:[a,b]\longrightarrow V$ and %$g:[a',b']\longrightarrow V$ where $(V,d)$ is a metric space,
%\begin{equation}
%    d_{F}(f,g)=\inf_{\alpha,\beta}\max_{t\in[0,1]}\{d(f(\alpha(t)),g((\beta(t))))\},
%\end{equation}
%where $\alpha$, $\beta$ are non-decreasing function from $[0,1]$ onto $[a,b]$ and $[a',b']$.

%We aim to use the kernels to generate high-dimensional projections that encode the influence each observed trajectory holds on predictions. 

We project each observed trajectory with DF-kernel onto a set of representative trajectories. We obtain $\bm{\varphi}_{n} \in \mathcal{R}^{M_{\bm{\xi}}}$, a vector of projections from ${\bm{\xi}_{n}}$ onto $\{\hat{\bm{\xi}_1},\ldots, \hat{\bm{\xi}_{M_{\xi}}}\}$. A set of $M_{\xi}$ trajectories, $\{\hat{\bm{\xi}_{1}},\ldots,\hat{\bm{\xi}_{M_{\xi}}}\}$, are selected from the set of all observed input trajectories. We refer to the selected trajectories as \textit{representative trajectories}. An alternative view of this process is placing basis functions over representative trajectories. The corresponding high-dimensional features over all $N$ observations are given by, 

%%Like many other distance substitute kernels, evaluating the DF-kernel function between all the data points is not guaranteed to have a positive semi-definite (psd) matrix. However, in this work we only operate on the feature maps, and not on a kernel matrix. As the inner product of the features guarantee a positive semi-definite matrix, the kernel matrix does not correspond to the inner product of feature vectors. The vector of features, $\bm{\varphi}$, can simply be thought of as a set of basis functions placed at the set of representing trajectories. The corresponding high-dimensional features over all $N$ observations is given by,

\begin{equation}
    K_{N\times M_{\xi}}=
    \begin{bmatrix}
    \bm{\varphi}_{1}^{T}\\
    \vdots\\
    \bm{\varphi}_{N}^{T}
    \end{bmatrix}
    =
    \begin{bmatrix} 
    k_{DF}(\bm{\xi}_1,\hat{\bm{\xi}_1}) & \ldots & k_{DF}(\bm{\xi}_1,\hat{\bm{\xi}_{M_{\xi}}}) \\
    \vdots & \ddots & \vdots\\
    k_{DF}(\bm{\xi}_N,\hat{\bm{\xi}_1}), & \ldots, & k_{DF}(\bm{\xi}_N,\hat{\bm{\xi}_{M_{\xi}}}).
    \end{bmatrix} 
\end{equation}

We later input the projection features to a simple neural network model, and do not operate directly on the Gram matrix. This can be viewed as learning combinations of fixed basis functions, similar to sparse Gaussian process (GP) regression \citep{SparseGP}. Selecting good representative trajectories can be done in a manner similar to selecting inducing points for the Nystr\"{o}m Method \citep{Nystrom,NystromA1,NystromA2,NystromA3} in sparse GPs. Even though randomly selecting a subset of trajectories from the observed trajectories is sufficient, we outline a quick and simple sampling scheme, similar to the leverage score sampling method \cite{NystromA1}. Provided a square matrix of the discrete Fr\'echet distances between all trajectories in a dataset of observation, $D_{N\times N}$, sort the columns of the matrix by its $L_{2}$ norm, and select every $i^{th}$ column, with $i$ being a fixed stepsize. The corresponding trajectory of each column selected is added to the representative set. The intuition is that almost identical trajectories would likely be sorted adjacent to one another. Hence, our heuristic discourages selecting multiple almost identical representative trajectories, and encourages selecting a more diverse set of representations. 

Though not explored deeply in this work, projecting input trajectories to a fixed set of representative trajectories may also be exploited to efficiently condition on high-dimensional trajectories. Challenges can arise from the "vastness" of space trajectories can lie in. If trajectories in high-dimensional space belong to only a few groupings, and in practice only occupy a limited volume in high-dimensional space, inputs may be adequately represented by a not-too-large representative set. 

The projected feature vectors generated are representations of our discrete input observations, whereas continuous output trajectories sampled from KTMs are in concise functional forms. Details for constructing functional trajectories are described in the next subsection.

\subsection{Constructing Continuous Functional Trajectories}\label{functionalTraj}

The conversion of target trajectories from ordered sets of coordinates to parameterised functions can be viewed as finding a concise low-dimensional representation of discrete trajectories. We assume that each output trajectory comprises independent functions, $x(t)$ and $y(t)$, that model the $x, y$ coordinates of the trajectory over time $t$. $x(t)$ and $y(t)$ give coordinates relative to the last waypoint coordinate of the queried discrete trajectory. A target trajectory recorded from time $T'$ to $T$, $\bm{\xi^{Tar}}=\big\{(x_t,y_t)\}^{T}_{t=T'}$, is represented as weighted sums of projections to square exponential basis functions placed at fixed times, $\bm{\Xi}(\cdot)=(\bm{w_x}^{T}\bm{\phi}(\cdot),\bm{w_y}^{T}\bm{\phi}(\cdot))$, where $\bm{\phi}(\cdot)$ represents the features, and $\bm{w_x}$, $\bm{w_y}$ are weights. Squared exponential basis functions are smooth and often used as a least informative default, though we are not restricted to using squared exponential bases. The weights parameters are found by solving kernel ridge regression problems with constraint $t=0$:

\begin{minipage}{0.49\textwidth}
\begin{subequations}
\begin{alignat}{2}
    & \min_{\bm{w_x}}&  & \sum_{n=1}^{T-T'}\big(x_n-\bm{w_x}^{T}\bm{\phi}(t_n) \big)^2 + \lambda_1||\bm{w_x}||^{2}\label{eq:optProb}\\
    &\text{s.t.} &   & \bm{w_x}^{T}\bm{\phi}(0)=0\label{con1}\
\end{alignat}
\end{subequations}
\end{minipage}
\begin{minipage}{0.49\textwidth}
\begin{subequations}
\begin{alignat}{2}
    &\min_{\bm{w_y}}&  & \sum_{n=1}^{T-T'}\big(x_n-\bm{w_y}^{T}\bm{\phi}(t_n) \big)^2 + \lambda_1||\bm{w_y}||^{2}\label{eq:optProb2}\\
    &\text{s.t.} &   & \bm{w_y}^{T}\bm{\phi}(0)=0\label{con2}\
\end{alignat}
\end{subequations}
\end{minipage}

where $\lambda_1$ is a regularisation coefficient, and $\bm{\phi(\cdot)}$ is a feature map defined by,
\begin{equation}
\bm{\phi}(t)=[k(t,\hat{t}_1),\ldots,k(t,\hat{t}_{M_t})]=\Bigg[\exp\Big(-\frac{||\hat{t}_1-t||^2}{2\ell_{t}}\Big),\ldots,\exp\Big(-\frac{||\hat{t}_{M_t}-t||^2}{2\ell_{t}}\Big)\Bigg]
\end{equation}
where $\hat{t_1},\ldots,\hat{t_{M_t}}$ is a set of $M_t$ fixed points in time. We refer to these points as \textit{inducing points}, and center the basis functions on them. $\ell_{t}$ is a length scale of the square exponential bases. Note that $\bm{\phi}$, projects to inducing points in time, and $\bm{\varphi}$, projects to representative trajectories. By including equations \ref{con1} and \ref{con2} as squared penalty terms, with penalty coefficient $\lambda_2$, to equations \ref{eq:optProb} and \ref{eq:optProb2}, and equating derivatives to zero gives the solution to the minimisation problems,
\begin{equation}
\begin{aligned}
    \bm{w_x}&=\Big(\lambda_1\bm{I}+\lambda_2\bm{\phi}(0)^{T}\bm{\phi}(0)+\sum_{n=1}^{N}\bm{\phi}(t_n)^{T}\bm{\phi}(t_n)\Big)^{-1}\Big(\sum_{n=1}^{N}{x_n\bm{\phi}(t_n)}\Big),\\
    \bm{w_y}&=\Big(\lambda_1\bm{I}+\lambda_2\bm{\phi}(0)^{T}\bm{\phi}(0)+\sum_{n=1}^{N}\bm{\phi}(t_n)^{T}\bm{\phi}(t_n)\Big)^{-1}\Big(\sum_{n=1}^{N}{y_n\bm{\phi}(t_n)}\Big)\label{eq:SolveW}.
\end{aligned}
\end{equation}
We can solve the minimisation problem to obtain vector of weights, $\bm{w_x}$ and $\bm{w_y}$, that parameterise the function $x(t)$ and $y(t)$ respectively. In this work, we define the same set of inducing points for $x(t)$ and $y(t)$, so both $\bm{w_x}$ and $\bm{w_y}$ are of dimensionality $M_t$, as there is a weight for each basis.

\subsection{Learning a Mixture of Stochastic Processes}\label{Learning Mixture}

We extend our functional representation of trajectories to stochastic processes, akin to distributions over functions. To model stochastic processes $\{x_t\}_t$ and $\{y_t\}_t$, we fit distributions over the weight parameters of $x(t)$ and $y(t)$. Namely, we wish to find the probability distribution, $p(\bm{w}|\bm{\xi}^{*},\mathcal{D})$, where $\bm{w}$ is a vector containing both $\bm{w_x}$ and $\bm{w_y}$, and $\bm{\xi}^{*}$ is a queried trajectory. We consider the concatenation of vectors $\bm{w_x}$ and $\bm{w_y}$, $\bm{w}$ which has $2M_t$ elements. To permit multiple modes over the mean function, assume $\{x_t\}_t$ and $\{y_t\}_t$ can be expressed as a linear sum of $R$ individual stochastic processes, which we shall call components. We can express $p(\bm{w}|\bm{\xi}^{*},\mathcal{D})$ as a linear sum with mixture coefficients $\alpha_{r}[\bm{\varphi}]$, where $\sum_{r=1}^{R}\alpha_{r}[\bm{\varphi}]=1$. Each $\alpha_{r}[\bm{\varphi}^{*}]$ is a function on $\bm{\varphi}^{*}$, the projections of $\bm{\xi}^{*}$ via the DF-kernel, detailed in subsection \ref{Subsection:genfeat}. Defining the shorthand $\alpha_{r}:=\alpha_{r}[\bm{\varphi}^{*}]$, we have, 
\begin{equation}\label{MDNeqn}
    p\big(\bm{w}|\bm{\xi}^{*},\mathcal{D}\big) = p(\bm{w}|\bm{\varphi}^{*})=\sum_{r=1}^{R}\alpha_{r}p_{r}(\bm{w}|\bm{\varphi}^{*}).
\end{equation}
In this work, we approximate the probability distribution of each element of $\bm{w}$ in each component, given a queried trajectory $\bm{\xi}^{*}$, to be independent Gaussian distributions. The mean, $\mu_{r,m}[\varphi^{*}]$, and standard deviations, $\sigma_{r,m}[\varphi^{*}]$, of the $m^{th}$ weight of the $r^{th}$ component are functions of $\bm{\varphi}^{*}$. For brevity, we use the shorthand $\mu_{r,m}:=\mu_{r,m}[\varphi^{*}]$ and $\sigma_{r,m}:=\sigma_{r,m}[\varphi^{*}]$. For the $m^{th}$ weight of the $r^{th}$ component, we have $p_r(w_m|\bm{\varphi}^{*})=\mathcal{N}(\mu_{r,m},\sigma_{r,m}^{2})$. Assuming weights are independent, the conditional probability over the vector of weights, of each component $r$, is $p_r(\bm{w}|\bm{\varphi}^{*})=\prod_{m=1}^{2M}\mathcal{N}(\mu_{r,m},\sigma_{r,m}^{2})$. We subsequently derive a loss function to learn $\mu_{r,m}$, $\sigma_{r,m}$, and $\alpha_r$, for all $r$ and $m$.

Let us consider the set of $N$ observations of input and target trajectories, $\mathcal{D}=\{(\bm{\xi}^{\text{Obs}},\bm{\xi}^{\text{Tar}})_n\}_{n=1}^{N}$. At the $n^{th}$ observation, $\bm{\xi^{\text{Obs}}_{n}}$ is projected using the DF-kernel to obtain high-dimensional projections, $\bm{\varphi_n}$. Weights, $\bm{w_{n}}$, that parameterise
$\bm{\Xi}^{\text{Tar}_{n}}$, continuous representations of discrete target trajectories, are then found by evaluating equation \ref{eq:SolveW}. Assuming that observations are independent and identically distributed, we can write the conditional density as,
\begin{equation}\label{eq:conditionalDensity}
    p(\{\bm{w_n}\}^{N}_{n=1}|\{\bm{\varphi_{n}}\}^{N}_{n=1})=\prod_{n=1}^{N}p(\bm{w_n}|\bm{\varphi_n})
    =\prod_{n=1}^{N}\sum_{r=1}^{R}\alpha_{r}[\bm{\varphi_n}]\prod_{m=1}^{2M_t}\mathcal{N}(\mu_{r,m},\sigma_{r,m}^{2})
\end{equation}
Fitting the conditional probabilities over weight parameters can be done by maximising \ref{eq:conditionalDensity}. We define the loss function as, 
\begin{align}
    \mathcal{L}&=-\log\big\{p(\{\bm{w_n}\}^{N}_{n=1}|\{\bm{\varphi_{n}}\}^{N}_{n=1})\big\}\\
    %&=-\sum_{n=1}^{N}\log\bigg\{\sum_{r=1}^{R}\alpha_{r}\prod_{m=1}^{2M_t} \frac{1}{\sqrt{2\pi\sigma^{2}_{r,m}}}\exp\bigg\{ -\frac{(w_{n,m}-\mu_{r,m})^2}{2\sigma^{2}_{r,m}}\bigg\}\bigg\}\\
    &=-\sum_{n=1}^{N}\log\Big\{\sum_{r=1}^{R}\exp\Big[ \log(\alpha_r)-2M\log(2\pi)+\sum_{m=1}^{2M}\log(\sigma_{r,m})-\sum_{m=1}^{2M}\frac{(w_{n,m}-\mu_{r,m})^2}{2\sigma^{2}_{r,m}}\Big]\Big\} 
\end{align}
Constraints $\sum_{r=1}^{R}\alpha_r=1$ can be enforced by applying a softmax activation function, $\alpha_r=\frac{\exp(z_{r}^{a})}{\sum_{r=1}^{R}\exp(z_{r}^{a})}$, where $z_{r}^{a}$ denotes the network outputs of $\alpha_r$. To enforce $\sigma_{r,m}\geq 0$, an exponential activation function, $
    \sigma_{r,m}=\exp(z^{\sigma}_{r,m})$, is applied to the network outputs corresponding to standard deviation. By utilising the expressiveness of our projection features, a simple mixture density network \citep{Bishop94mixturedensity,MDNABrandoMasterThesis}, with a single hidden layer can then be used to learn the functions of parameters $\alpha_{r}[\bm{\varphi}]$, $\mu_{r,m}[\bm{\varphi}]$, $\sigma_{r,m}[\bm{\varphi}]$, by minimising our loss function via Stochastic Gradient Descent (SGD). 

\subsection{Conducting Inference and Obtaining Trajectory Realisations}\label{inferenceSubsection}
After learning the functions $\alpha_{r}[\bm{\varphi}]$, $\mu_{r,m}[\bm{\varphi}]$, and $\sigma_{r,m}[\bm{\varphi}]$ as described in subsection \ref{Learning Mixture}, we have $p\big(\bm{w}|\bm{\xi}^{*},\mathcal{D}\big)$ via equation \ref{MDNeqn}, and the assumption of independent Gaussian distributed weights. Given a vector of feature maps, $\bm{\phi}(t)$, to evaluate $p(\bm{\Xi}^{*}(t)|\bm{\xi}^{*},\mathcal{D}, \bm{\phi}(t))=\int p(\Xi^{*}(t)|\bm{\phi}(t), \bm{w})p(\bm{w}|\bm{\xi}^{*},\mathcal{D})d\bm{w}$, we have $p(\bm{\Xi}^{*}(t)|\bm{\phi}(t), \bm{w})=\mathcal{N}(\bm{w}^{T}\bm{\phi}(t),s^{2}\mathcal{I})$ \citep{BishopBook}, where $s$ denotes the standard deviation of the observation error. It is possible to estimate $s^{2}$ via $p(s^{2}|\mathcal{D})\propto p(\mathcal{D},s^2)=\int p(D|s^2,\bm{w})p(\bm{w})p(s^2)d\bm{w}$. Like \citep{BODeterministic1} and \citep{BODeterministic2}, in this work, we focus on the deterministic observation case, where $s=0$.
%$\bm{\Xi} \sim p(\Xi^{*}(t)|\bm{\phi(t)}, \bm{w})$ can be evaluated as $(\bm{w}_{x}^{T}\bm{\phi(\cdot)},\bm{w}_{y}^{T}\bm{\phi(\cdot)})$. 

The inference process to sample continuous trajectories $\bm{\Xi^{\text{out}}}$ is outlined in algorithm \ref{InferenceAl}. Under the assumption of deterministic observations, we evaluate $p(\bm{\Xi}^{*}|\bm{\xi}^{*},\mathcal{D}, \bm{\phi})=\int p(\bm{\Xi}^{*}|\bm{\phi}, \bm{w})p(\bm{w}|\bm{\xi}^{*},\mathcal{D})d\bm{w}$, by randomly sampling $p\big(\bm{w}|\bm{\xi}^{*},\mathcal{D}\big)$, and obtaining realisations of continuous trajectories $\bm{\Xi}^{out}(\cdot)\sim p(\bm{\Xi}^{*}(\cdot)|\bm{\phi(\cdot)}, \bm{w})$ by evaluating $(\bm{w}_{x}^{T}\bm{\phi(\cdot)},\bm{w}_{y}^{T}\bm{\phi(\cdot)})$. We can obtain a discrete trajectory $\bm{\xi}^{out}$ by querying $\bm{\Xi}^{out}(\cdot)$ at times, $\bm{t}=[t_{1}, \ldots, t_{n}]$, i.e. $\bm{\xi}
^{out}\leftarrow \bm{\Xi}^{out}(\bm{t})$. 

%we can carry out inference on an observed trajectory of interest, $\xi^{*}=\{x,y\}_{t=1}^{T^{*}}$, and sample realisations to obtain possible future trajectories. The process to perform inference is detailed in algorithm \ref{InferenceAl}. We require the DF-kernel with representative trajectories, $\bm{\varphi(\cdot)}=k_{DF}^{'}(\cdot,\bm{\hat{\xi}})$ and the rbf kernel stationed along time, $\bm{\phi(\cdot)}=k(\cdot,\bm{\hat{t}})$, with associated inducing points in time. A vector of time points where we wish to query the trajectory is given by $\bm{t}=[t_1,\ldots,t_{out}]$. Each continuous trajectory realised, $\bm{\Xi^{out}}$, is drawn from $p\big(\bm{\Xi^{out}}_{t \geq T'}|\bm{\xi}^{*}_{t=1,\ldots,T'},\mathcal{D}\big)$. By repeatedly sampling continuous trajectories, we can approximate the distribution over possible future trajectories.

\begin{algorithm}[h]
\SetKwInOut{Input}{input}
\SetKwInOut{Output}{output}
\caption{KTM Inference}\label{InferenceAl}

\Input{$\bm{\xi}^{*}$,$\alpha_{r}[\bm{\varphi}]$, $\mu_{r,m}[\bm{\varphi}]$, $\sigma_{r,m}[\bm{\varphi}]$, $\bm{\varphi(\cdot)}$, $\bm{\phi(\cdot)}$ }
\Output{Realised Continuous Trajectory,  $\bm{\Xi}^{\text{out}}(\cdot)$}
\BlankLine
\Begin{
$\bm{\varphi}^{*} \leftarrow \bm{\varphi(\bm{\xi}^{*})}$ \tcp{generate projections with DF-kernel}
Evaluate $\alpha_{r}[\bm{\varphi}^{*}]$, $\mu_{r,m}[\bm{\varphi}^{*}]$, $\sigma_{r,m}[\bm{\varphi}^{*}]$ \tcp{Find parameters of mixture of SP}
$p(\bm{w}|\bm{\varphi}^{*})\leftarrow \sum_{r=1}^{R}\alpha_{r}p_{r}(\bm{w}|\bm{\varphi}^{*})$\\
$\bm{w} \sim p(\bm{w}|\bm{\varphi}^{*})$ \tcp{Sample $p(\bm{w}|\bm{\varphi}^{*})$}
$\bm{\Xi}^{out}(\cdot)\leftarrow (\bm{w}_{x}^{T}\bm{\phi(\cdot)},\bm{w}_{y}^{T}\bm{\phi(\cdot)})$ \tcp{retrieve continuous trajectory}
%$\Phi_{t}\leftarrow[\bm{\phi(t_1)},\ldots,\bm{\phi(t_{out})}]^{T}$ \tcp{Project points of interest to inducing points}
%$\bm{\xi_{out}}\leftarrow \bm{w}^{T} \Phi_{t}$
}\end{algorithm}

\section{Experiments and Discussions}
We wish to highlight the benefits KTMs bring. In particular: (1) map-awareness; (2) trajectory history awareness; (3) multi-modal probabilistic predictions, with continuous trajectory realisations. 

\subsection{Experimental Setup} \label{setup}
We run experiments on both simulated and real-life trajectory datasets, including:

\begin{enumerate}
    \item Simulated dataset (S): Simulated trajectories of pedestrians crossing a road, similar to the simulated datasets used in \citep{DirectionalGridMaps}
    \item Edinburgh dataset \citep{Edenburgh} (E): Pedestrian trajectories in the real-world on September $24^{th}$
    \item Lankershim dataset \citep{Lankershim} (L): Subset of valid vehicle trajectories in the region between x-coordinates $-100$m$ \sim 100$m and y-coordinates $250$m$\sim 500$m
\end{enumerate}

In experiments, each whole trajectory is segmented with an input-target ratio of 1:3, 1:1, or 3:1. The subset of the Lankershim dataset contains 6580 pairs of trajectories; Edinburgh 5972; simulated 600. We use $R=4$ mixture components, and length scale $\ell_{DF}=100$, for the DF-kernel, and $\ell_{t}=10$ for the square exponential bases. Bases are centered evenly at 2.5 time step intervals for the Edinburgh dataset and 5 for the simulated and Lankershim datasets. Half the trajectories are used as representative trajectories. To adequately evaluate the ability of KTMs, we ensure representative trajectories are not included in testing. We randomly select 20\% of trajectories outside of the representative set as test examples, or $10\%$ of the total. To account for stationary vehicles, for the Lankershim dataset, we only evaluate trajectories that move more than 20m in 20 time steps. All values reported in metres. We train for 80 epochs, then evaluate on the test set. Inference can be conducted efficiently, with an average time below 0.2 sec for predicting mixture of processes, on all of our experiments with a standard desktop. Experiments are repeated 5 times, each with randomly selected test examples. We give quantitative results on the following realised trajectories from the output:

\begin{enumerate}
    \item KTM-Weighted Average (KTM-W): A linear combination of the mean of each mixture component, weighted by the mixture coefficient;
    \item KTM-Closest (KTM-C): The mean trajectory of the mixture component that is the closest to ground truth. Selecting the trajectory in this manner assumes the decision of which option, out of the four possible trajectories to take, is made correctly. This allows us to evaluate the quality of the predicted trajectory, without taking into account of the quality of decision-making;
    \item Constant Velocity (CV): The trajectory is generated by a model that the velocity remains constant beyond the observations;
    \item Directional Grid Maps (DGM): Directional grid map \citep{DirectionalGridMaps} is a recent method capable of producing directional predictions. We conduct forward sampling on a DGM, with a step size equal to that of the last observed step.
\end{enumerate}

The metrics used to evaluate our trajectories are:  (1) Euclidean distance (ED) between the end points of predicted and ground truth trajectories; 
(2) Discrete Fr\'echet distance (DF) \citep{DiscreteFrechet, TrajectoryDistances} between predicted and ground truth trajectories. Continuous trajectories are discretised for comparison. 

\subsection{Map-Awareness}
Kernel Trajectory Maps learn to predict trajectories from a dataset of observed trajectories, which contain rich information about the structure of the environment, such as obstacles and designated paths. Methods that learn from a set of observed trajectories are intrinsically \emph{map-aware} \cite{HumanMotionSurvey}, and can account for environment geometry. Dynamics based models are often map-unaware, and are not able to anticipate a future changes in direction due to environmental factors, such as a turning road.

An example of map-awareness is demonstrated in figure \ref{fig:Pred_turn}. We sample realisations from the predicted mixture of a KTM, and compare it against the constant velocity (CV) model and ground truth trajectory. The sharp turn the ground truth trajectory takes is due to the road structure in the dataset, and there is little indication from the behaviour of the observed trajectory. The turn is not captured by the CV model, but is captured by the KTM.

Table \ref{ResultsTable1} contains the quantitative results of methods described in subsection \ref{setup}. We predict future trajectories over a horizon of 20 time steps. We see that map-aware methods, such as KTM and DGM, tend to outperform the CV model. Notably the CV model performs strongly for the Lankershim dataset, outperforming all but the KTM-C method, due to vehicle trajectories in that dataset being approximately constant velocity over small distances. The Edinburgh dataset contains pedestrian motion trajectories which are much more unstructured and unrestricted. Thus, KTM-C and KTM-W perform significantly stronger than the CV model.  

\begin{minipage}[t]{0.30\textwidth}
\strut\vspace*{-\baselineskip}\newline
    \centering
      \includegraphics[width=0.95\textwidth]{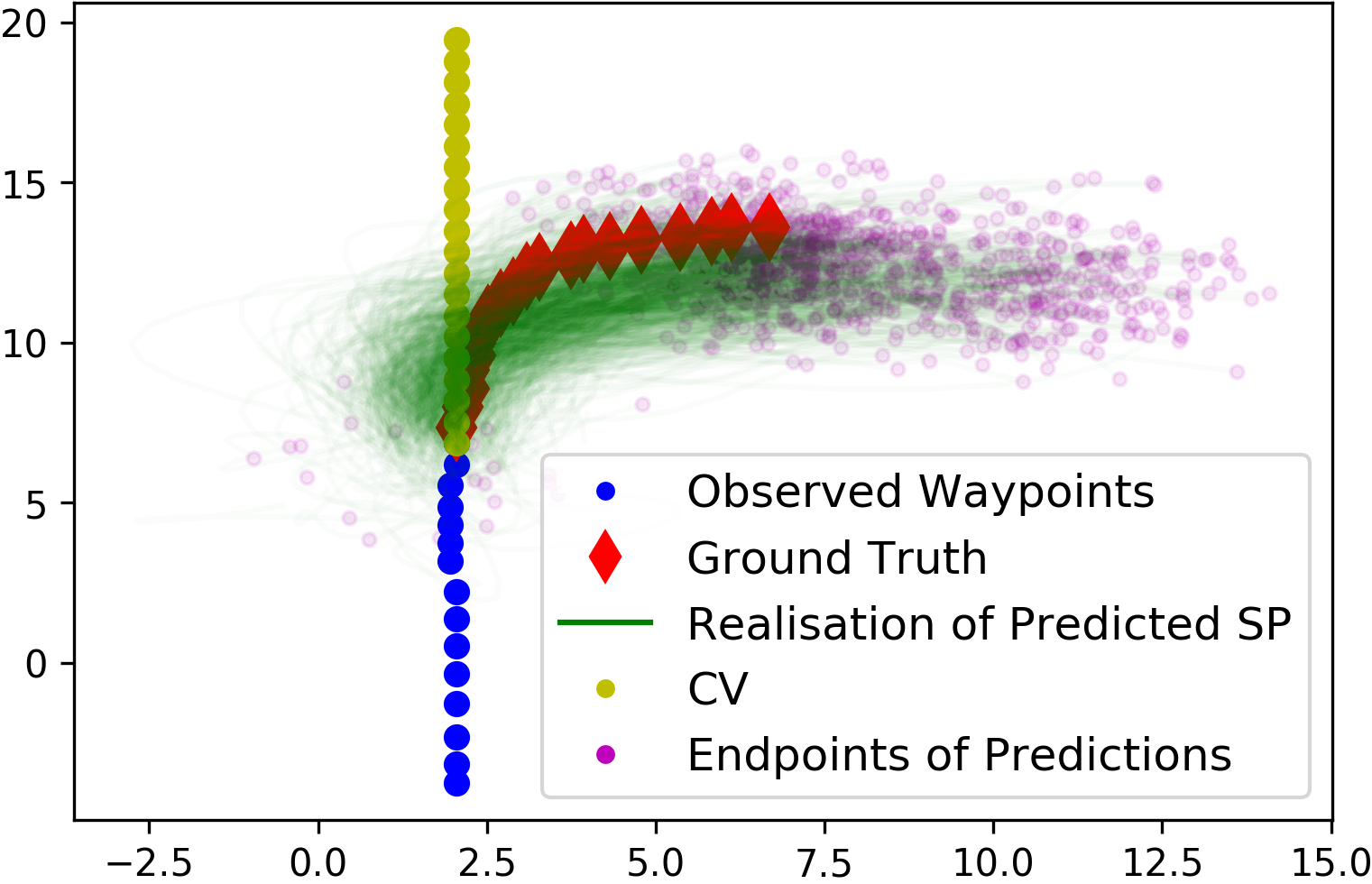}
      %\vspace*{-3mm}
       \captionof{figure}{1000 sampled trajectories from output mixture (green with magenta endpoints) anticipate the turn, as shown by the ground truth (red). There is little indication of the turn from observed waypoints (blue). The CV (yellow) model does not.} \label{fig:Pred_turn}
\end{minipage}
\hspace{0.03\linewidth}
\begin{minipage}[t]{0.65\textwidth}
\strut\vspace*{-\baselineskip}\newline
\def\arraystretch{0.8}
\setlength{\tabcolsep}{4pt}
\centering
    \begin{tabular}{cc|rrrr}
    \hline
\multicolumn{1}{l}{}                           & \multicolumn{1}{l}{}   & KTM-C   & KTM-W    & CV       & DGM \\ \hline
\multirow{2}{*}{(S)} & ED & 1.3$\pm$0.1 & 1.8$\pm$0.2  & 6.5$\pm$0.3  & 4.4$\pm$0.1                 \\
\multicolumn{1}{l}{}                           & DF & 1.4$\pm$0.1 & 1.9$\pm$0.1  & 6.3$\pm$0.3  & 4.4$\pm$0.1                 \\
\multirow{2}{*}{(E)}                     & ED                     & 0.7$\pm$0.1 & 0.9$\pm$0.1  & 1.4$\pm$0.1  & 1.1$\pm$0.1                 \\
                                               & DF                     & 0.8$\pm$0.1 & 0.9$\pm$0.1  & 1.4$\pm$0.1  & 1.1$\pm$0.1                 \\
\multirow{2}{*}{(L)}                    & ED                     & 5.8$\pm$0.3 & 11.5$\pm$0.2 & 11.3$\pm$0.2 & 11.4$\pm$0.2                \\
                                               & DF                     & 6.3$\pm$0.3 & 11.5$\pm$0.5 & 10.7$\pm$0.2 & 11.4$\pm$0.2\\
                                               \hline
\end{tabular} 
   \captionof{table}{The performance of KTMs compared to a baseline CV model and an map-aware DGM model \citep{DirectionalGridMaps}, on the Simulated dataset (S), the Edinburgh dataset (E), and the Lankershim dataset (L). We see that KTM-C outperforms the other methods, while KTM-W also gives a strong performance. KTMs benefit from map and trajectory history awareness. Note that the CV model performs well on the Lankershim dataset (L) due to the vehicle trajectories being approximately constant velocity of relatively short time horizons. Results given in meters. }\label{ResultsTable1}

\end{minipage}
\subsection{Trajectory History Awareness}
\begin{figure}
    \centering
    \begin{subfigure}
     \centering
     \includegraphics[width=0.25\textwidth]{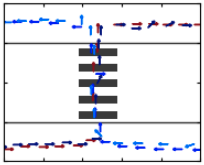}
    \end{subfigure}
    \begin{subfigure}
     \centering
     \includegraphics[width=0.34\textwidth]{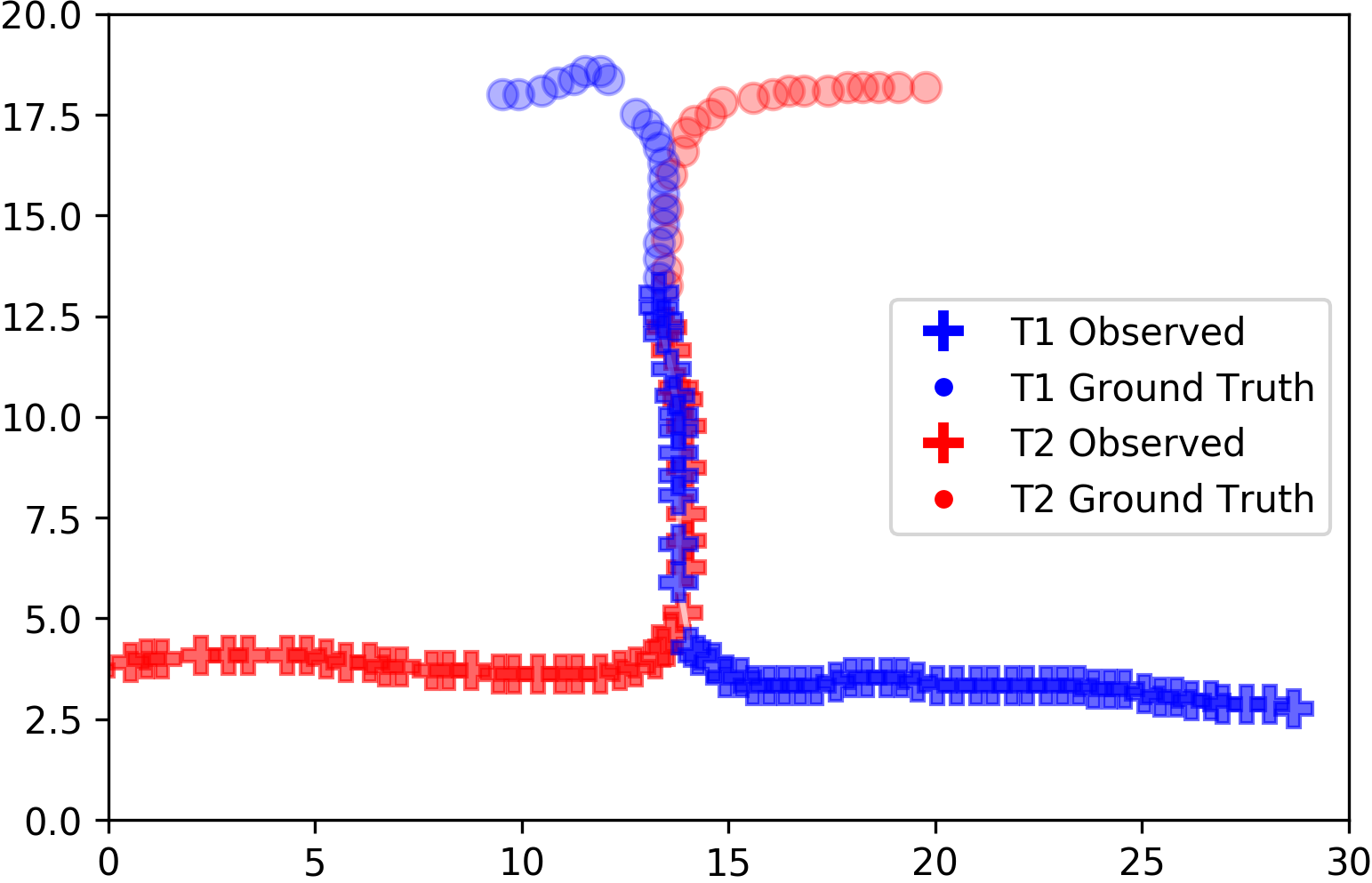}
    \end{subfigure}
    \begin{subfigure}
     \centering
     \includegraphics[width=0.34\textwidth]{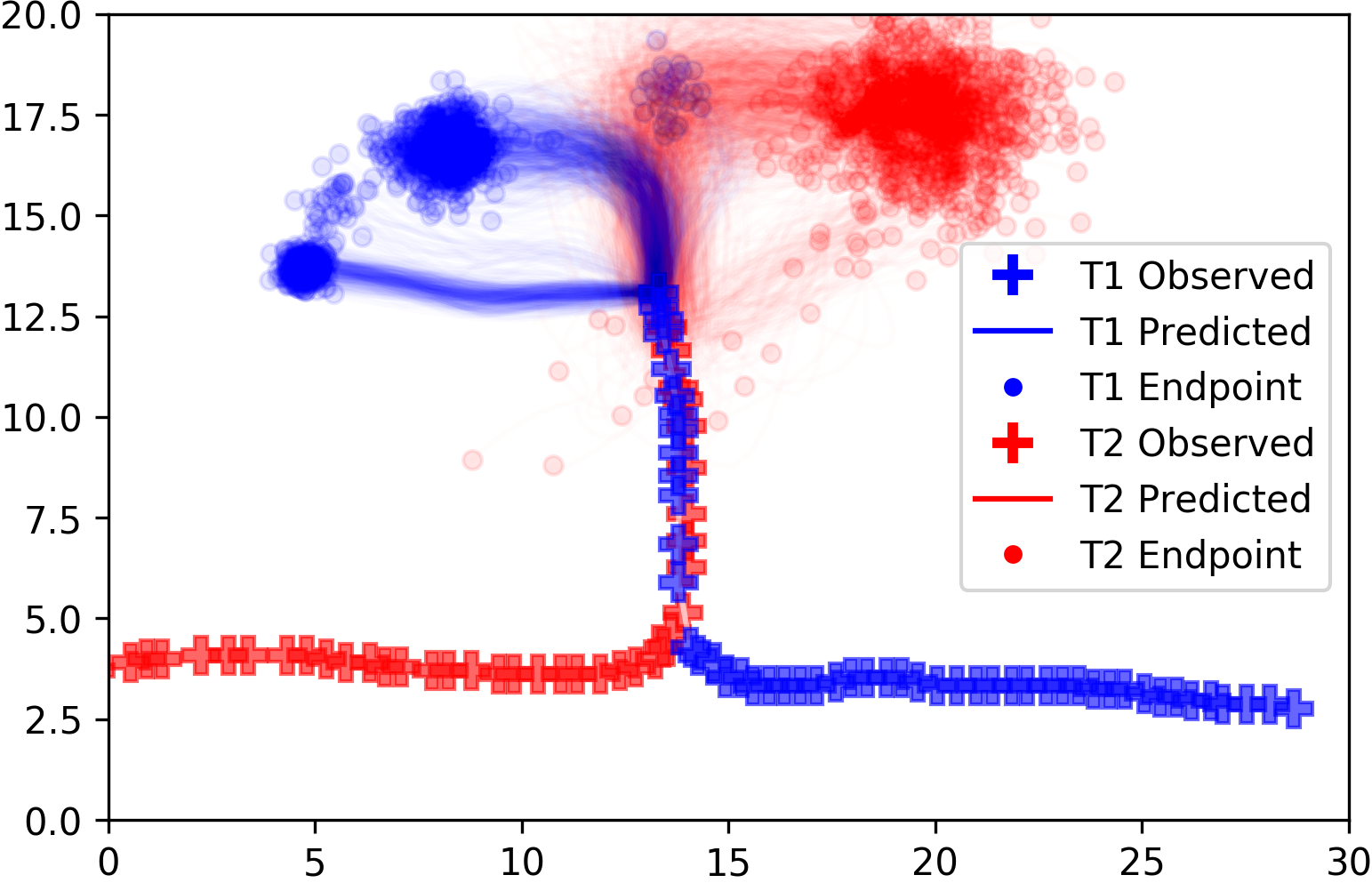}
    \end{subfigure}
    \vspace*{-3mm}
    \caption{Examples of simulated trajectories (left). All trajectories starting at the lower left terminate at the upper right, and those starting at the lower right terminate at the upper left. The ground truth of the two trajectories, one starting at the lower left, the other at the lower right, are shown (center). Though the latest waypoints of both are similar, the KTM predictions are visibly different.}\label{historyPlot}

    %\caption{Sampled trajectories (left) from a KTM trained on simulated trajectories. Ground truth (center), and examples of the simulated trajectories (right) are shown. Though the last coordinate of observed $T1$ and $T2$ are similar, the predictions are visibly different.}\label{historyPlot}
\end{figure}

Recent attempts to encode multi-modal directional distributions in a map \citep{DirectionalGridMaps,Flow,KernelEmbeddingDirections} largely condition only on the most recent coordinate, and are unable to utilise the full trajectory history of the object. KTMs are trajectory history aware, as demonstrated by trajectories sampled from a KTM trained on the simulated dataset, shown in figure \ref{historyPlot}. The predicted trajectories sampled vary significantly, though the positions of the last observed location are similar. Methods that condition solely on the most recent coordinate, can not differentiate between the two observed trajectories. The latter portion of the observed trajectories are similar, but with dissimilar early portions. By exploiting DF-kernels, KTMs give predictions conditioned on the entire trajectory. Although directional flow methods, such as DGM \citep{DirectionalGridMaps}, are able to capture the general movement directions of dynamic objects, trajectories can only be obtained by making the Markov assumption and forward sampling. This process is sensitive to errors, and recursive behaviour can also arise. For example, a prediction at A points to B, which in turn may give a prediction pointing back to A. KTMs allow for realisations of entire trajectories, without forward sampling or making Markovian assumptions. The trajectory history awareness of KTMs explain the strong performance of KTMs relative to the DGM method, specifically on the simulated dataset, as shown in table \ref{ResultsTable1}.
\begin{figure}
    \centering
    \begin{subfigure}
     \centering
     \includegraphics[width=0.32\textwidth]{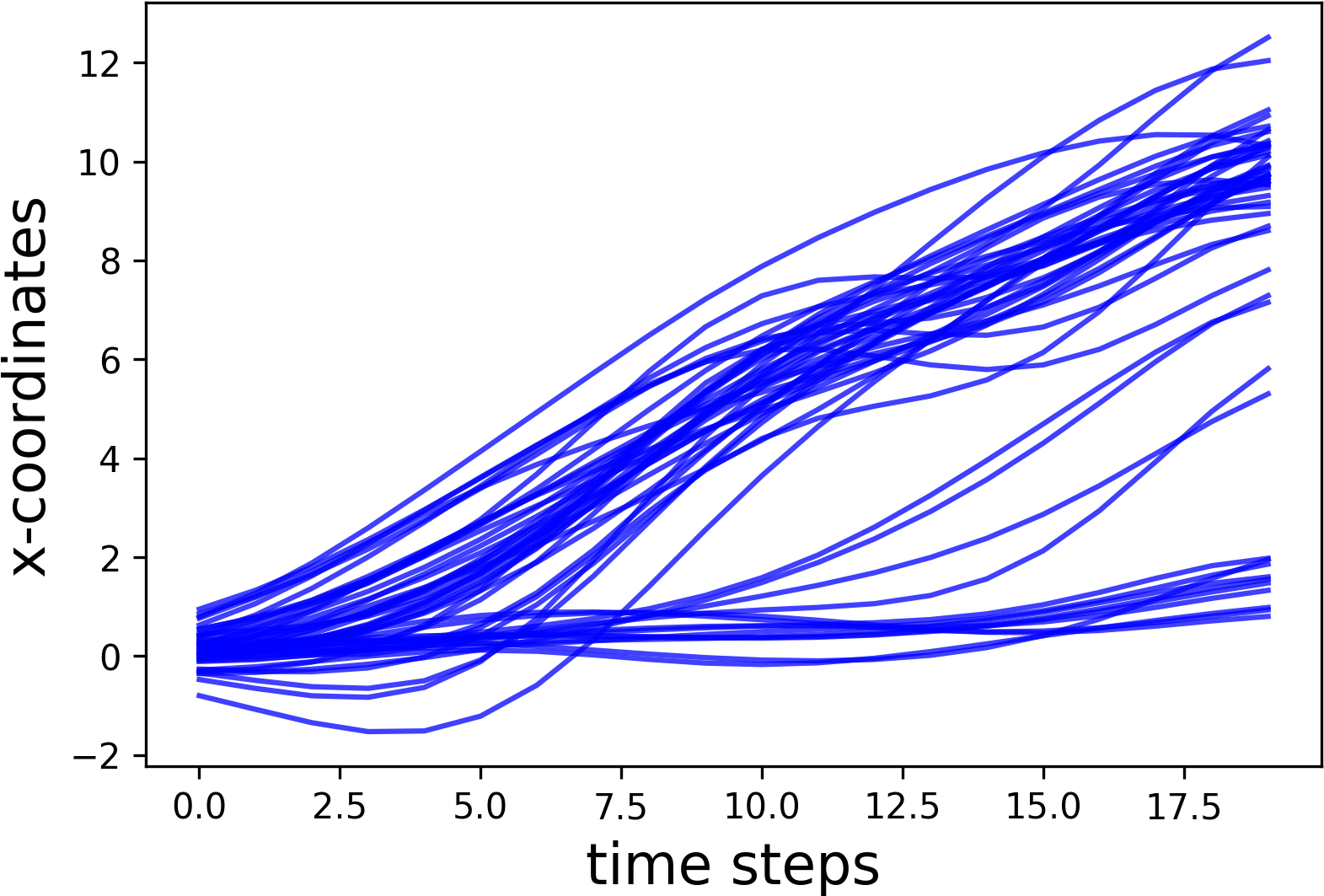}
    \end{subfigure}
    \begin{subfigure}
     \centering
     \includegraphics[width=0.32\textwidth]{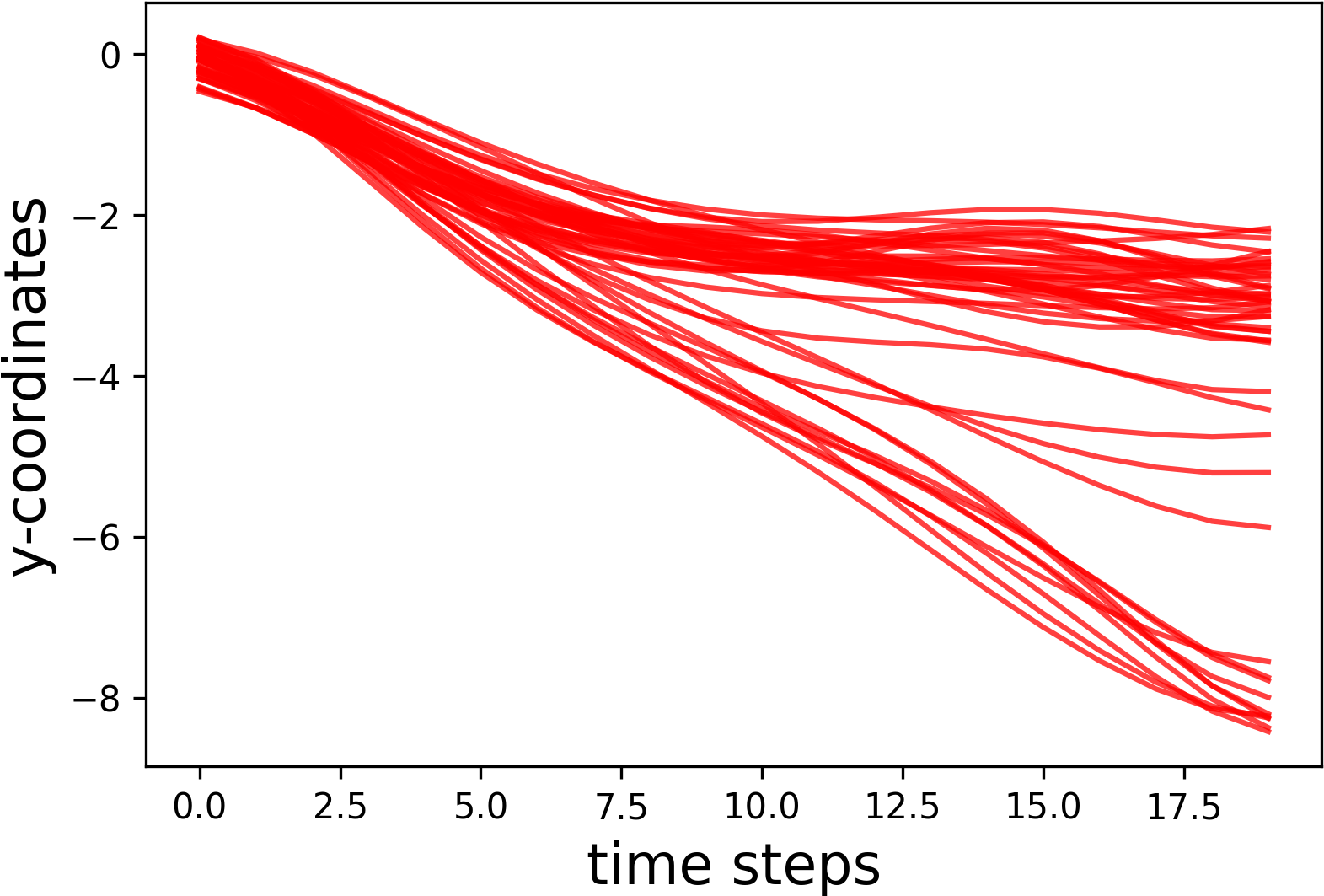}
    \end{subfigure}
    \begin{subfigure}
     \centering
     \includegraphics[width=0.32\textwidth]{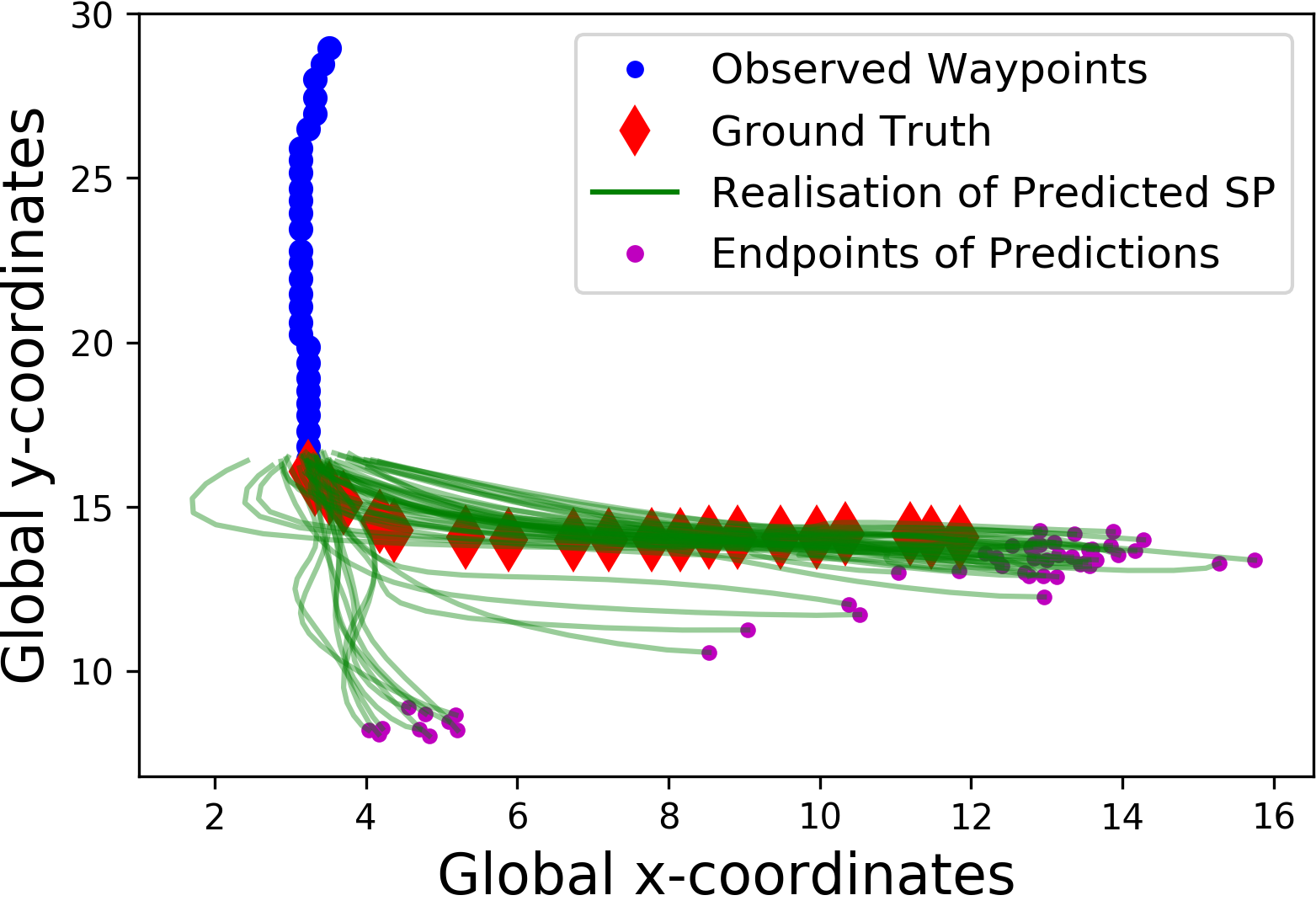}
    \end{subfigure}
    \vspace*{-3mm}
    \caption{50 realisations of $x(t)$ and $y(t)$ (left and center respectively), and the corresponding predicted trajectories (right). $x(t)$ and $y(t)$ give coordinates relative to the last observed coordinate.}\label{outputPlots}
\end{figure}
\subsection{Multi-modal Probabilistic Continuous Outputs}
KTMs output mixtures of stochastic processes, corresponding to multi-modal distributions over functions. This provides us with information about groups of possible future trajectories with associated uncertainty. Figure \ref{outputPlots} illustrates sampling functions from the outputted mixtures. The left and center plots show realisations of the functions $x(t)$ and $y(t)$, and the right plot shows the corresponding trajectory. There is clear multi-modality in the distribution over future trajectories. 

A major benefit of KTMs is that realisations of the output are continuous functional trajectories. These are smooth and continuous, and do not commit to an \emph{a priori} resolution. We can query any time value to retrieve predicted coordinates at the given time point. The functional representation with square exponential bases is inherently smooth, allowing us to operate on the derivatives of displacement. This property permits us to constrain certain velocity, acceleration, or jerk values.   
\section{Conclusion}
In this paper, we introduce Kernel Trajectory Maps (KTM), a novel multi-modal probabilistic motion prediction method. KTMs are map-aware and condition on the whole observed trajectory. By projecting on a set of representative trajectories using expressive DF-kernels, we can use a simple single hidden layer mixture density network to arrive at a mixture of stochastic processes, equivalent to a multi-modal distribution over future trajectories. Each realisation of the mixture is a continuous trajectory, and can be queried at any time resolution. We recover whole trajectories without resorting to forward sampling coordinates. Empirical results show the awareness of the map and trajectory history improves performance when compared to a CV and map-aware, but not trajectory history aware, DGM model. The multi-modal and probabilistic properties of KTMs are also apparent from the experimental results. 
Future work will look into embedding social dynamics, and interaction between multiple predicted trajectories, into the KTM framework.
\section*{Acknowledgements}
The authors thank Rafael Oliveira and Philippe Morere for the fruitful discussions. W. Zhi is supported by a Commonwealth of Australia Research Training Program Scholarship.

\bibliography{Ref}
\end{document}